\documentclass[11pt, logo, onecolumn, copyright, colorlinks=true, allcolors=blue]{nvidiatechreport}
\usepackage{graphicx}
\usepackage{subcaption,ragged2e}
\usepackage[round]{natbib}
\usepackage[normalem]{ulem}
\usepackage[utf8]{inputenc}
\usepackage[T1]{fontenc}
\usepackage{url}
\usepackage{tikz}
\usepackage{enumitem}
\usetikzlibrary{positioning}
\usetikzlibrary{calc}
\usepackage{array}
\usepackage{booktabs}
\usepackage{wrapfig}
\usepackage{caption}
\usepackage{hyperref}
\usepackage{listings}
\usepackage{adjustbox}
\usepackage{footnote}
\usepackage{multirow}
\usepackage{amsmath}
\usepackage{amsfonts}
\usepackage{breakcites}
\usepackage{blindtext}
\usepackage{svg}
\usepackage[most]{tcolorbox}
\usepackage{tabularx}
\usepackage{graphicx} 
\usepackage{xcolor}
\usepackage{wrapfig}
\newcommand{\newparagraph}[1]{\noindent\textbf{#1\hspace{0.5em}}}

\newcommand{\HM}[1]{{\color{blue}{[}\textbf{Huizi: #1}{]}}}
\newcommand{\MX}[1]{{\color{purple}{[}\textbf{Meng: #1}{]}}}
\newcommand{\ST}[1]{{\color{brown}{[}\textbf{Sharath: #1}{]}}}
\newcommand{\TB}[1]{{\color{pink}{[}\textbf{Tijmen: #1}{]}}}

\title{Quantization-Aware Distillation for NVFP4 Inference Accuracy Recovery}
\author{Meng Xin, Sweta Priyadarshi, Jingyu Xin, Bilal Kartal, Aditya Vavre, Asma Kuriparambil Thekkumpate, Zijia Chen, Ameya Sunil Mahabaleshwarkar, Ido Shahaf, Akhiad Bercovich, Kinjal Patel, Suguna Varshini Velury, Chenjie Luo, Zhiyu Cheng, Jenny Chen, Chen-Han Yu, Wei Ping, Oleg Rybakov, Nima Tajbakhsh, Oluwatobi Olabiyi, Dusan Stosic, Di Wu, Song Han, Eric Chung, Sharath Turuvekere Sreenivas, Bryan Catanzaro, Yoshi Suhara, Tijmen Blankevoort, Huizi Mao\footnote{Contact: huizim@nvidia.com}}
\date{}

\begin{document}

\begin{abstract}

\textbf{Abstract.}
This technical report presents quantization-aware distillation (QAD) and our best practices for recovering accuracy of NVFP4-quantized large language models (LLMs) and vision-language models (VLMs). QAD distills a full-precision teacher model into a quantized student model using a KL divergence loss. While applying distillation to quantized models is not a new idea, we observe key advantages of QAD for today's LLMs: 1. It shows remarkable effectiveness and stability for models trained through multi-stage post-training pipelines, including supervised fine-tuning (SFT), reinforcement learning (RL), and model merging, where traditional quantization-aware training (QAT) suffers from engineering complexity and training instability; 2. It is robust to data quality and coverage, enabling accuracy recovery without full training data. We evaluate QAD across multiple post-trained models including AceReason Nemotron, Nemotron 3 Nano, Nemotron Nano V2, Nemotron Nano V2 VL (VLM), and Llama Nemotron Super v1, showing consistent recovery to near-BF16 accuracy.

\smallskip
\textbf{NVFP4 checkpoints} \href{https://huggingface.co/nvidia/NVIDIA-Nemotron-3-Nano-30B-A3B-NVFP4}{Nemotron 3 Nano 30B-A3B} \includegraphics[height=0.9em]{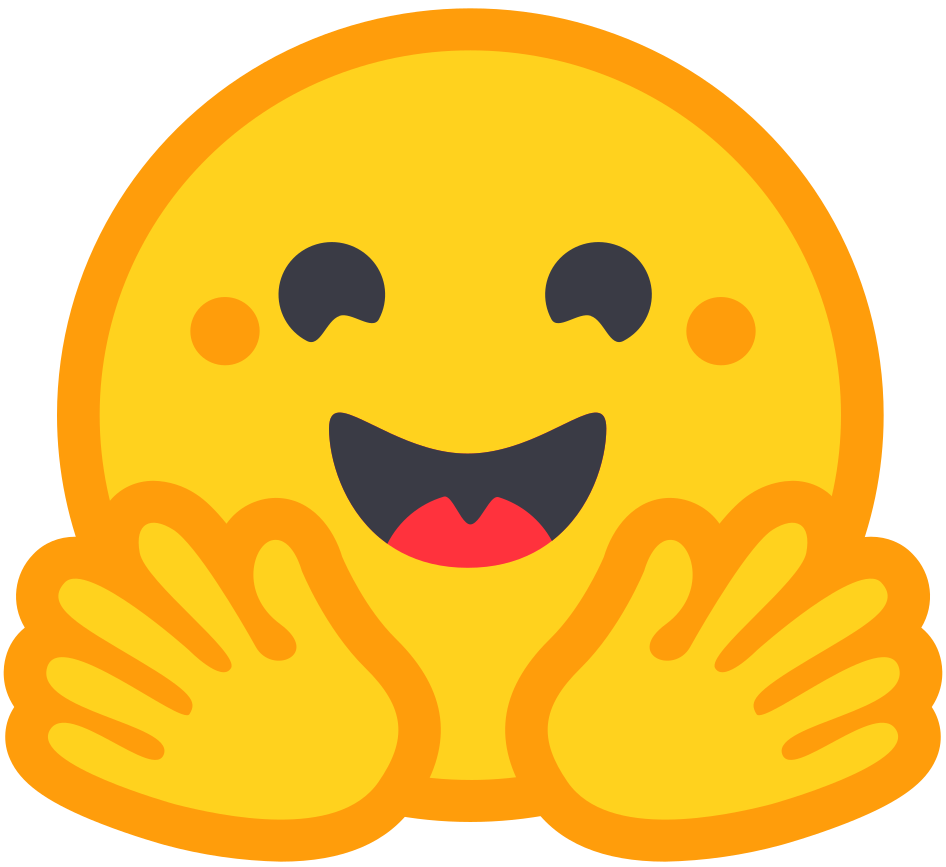},
\href{https://huggingface.co/nvidia/NVIDIA-Nemotron-Nano-9B-v2-NVFP4}{Nemotron Nano 9B V2} \includegraphics[height=0.9em]{assets/huggingface-color.png}, \href{https://huggingface.co/nvidia/NVIDIA-Nemotron-Nano-12B-v2-VL-NVFP4-QAD}{Nemotron Nano 12B V2 VL} \includegraphics[height=0.9em]{assets/huggingface-color.png}, \href{https://huggingface.co/nvidia/Llama-3.1-Nemotron-Nano-VL-8B-V1-FP4-QAD}{Llama 3.1 Nemotron Nano VL 8B} \includegraphics[height=0.9em]{assets/huggingface-color.png}

\textbf{QAD codes}: \href{https://github.com/NVIDIA/Model-Optimizer/tree/main/examples/llm_qad}{Megatron-LM version}, 
\href{https://github.com/NVIDIA/TensorRT-Model-Optimizer/tree/main/examples/nemo_run/qat}{NeMo version}, \href{https://github.com/NVIDIA/TensorRT-Model-Optimizer/tree/main/examples/llm_qat}{HuggingFace Transformers version} 


\end{abstract}

\maketitle

\section{Introduction}

The rapid expansion of large language models (LLMs) has increased the demand for more efficient numerical formats to lower computational cost, memory demand, and energy consumption during training and inference. 8-bit floating point formats (FP8 and MXFP8) have emerged as popular data types for accelerated training of LLMs~\citep{micikevicius2022fp8,deepseekv3}. Recent advances in narrow-precision hardware~\citep{nvidia2024blackwell} have positioned 4-bit floating point (FP4) as the next logical step~\citep{chmiel2025fp4,chen2025tetrajet,liu2023llmfp4,rouhani2023microscaling}, delivering a two- to three-fold boost in arithmetic performance and reducing memory usage by half compared to FP8.

NVFP4, which features a smaller block size (16 vs. 32 for MXFP4), FP8 scale factors (E4M3) for fine-grained scaling, and second-level FP32 scaling for a larger dynamic range, has demonstrated superior accuracy to INT4 and MXFP4 on many models~\citep{egiazarian2025bridging}. For very large LLMs, NVFP4 with post-training quantization (PTQ) shows decent accuracy on different benchmarks. However, for small LLMs, the accuracy drop from PTQ is often non-negligible.

There are many efforts to incorporate quantization into the training process. NVFP4 quantized training~\citep{nvidia2025nvfp4pretraining} has demonstrated promising convergence on pretraining tasks. With the primary goal of training speedup, models trained with NVFP4 are still evaluated in BF16. 

Quantization-aware training (QAT)~\citep{jacob2018quantization} is an effective training-based method for inference accuracy recovery, which has shown great results since the CNN era. Many QAT methods reuse the same pipelines and training objectives as the original high-precision models. However, this approach faces significant challenges for modern LLMs in real practices, such as:

\begin{enumerate}
    \item \textbf{Complex training pipelines}: Modern LLMs undergo multi-stage post-training~\citep{bakouch2025smollm3,bercovich2025llamanemotronefficientreasoningmodels}, such as SFT, RL, model merging, making it difficult to replicate the original training procedure.
    \item \textbf{Data availability and quality}: The original training data of open models may not be available, and public datasets are often of worse quality.
\end{enumerate}

This technical report evaluates quantization-aware distillation (QAD) for NVFP4 inference accuracy recovery on post-trained LLMs. QAD uses the original full-precision model as a teacher and trains the quantized model as a student using KL divergence loss rather than task-specific objectives. We demonstrate the following key findings:

\begin{enumerate}
    \item QAD aligns the quantized model to the high-precision model better than QAT.
    \item QAD works effectively for models trained through multi-stage post-training pipelines including SFT \& RL, demonstrating remarkable stability.
    \item QAD is robust to incomplete data coverage and can recover accuracy even with partial domain data, enabling cross-domain knowledge transfer.
\end{enumerate}

The report is organized as follows: Section~\ref{sec:nvfp4} provides a background on NVFP4 and quantization methods; Section~\ref{sec:qad} describes the QAD method and training setup and demonstrates key properties through comprehensive results across LLMs and VLMs; Section~\ref{sec:ablation} provides detailed analysis of design choices and the impact of training data.

\section{Background and Related Work}\label{sec:nvfp4}

\subsection{NVFP4 Format and Post-Training Quantization}


\textbf{NVFP4 Format}. NVFP4 is a 4-bit floating-point format designed for efficient training and inference on modern GPU architectures~\citep{nvidia2025nvfp4pretraining,nvidia2025nvfp4blog}. Compared to FP8, NVFP4 offers 2-3× higher arithmetic throughput and approximately 1.8× memory reduction. NVFP4 extends the MXFP4 format~\citep{rouhani2023microscaling} with a smaller block size (from 32 to 16) and two-level scaling (per-block E4M3 scales plus per-tensor FP32 scale). The smaller block size enables more localized adaptation to data distributions, while E4M3 scales provide non-power-of-two scaling factors for lower quantization error, and the second-level FP32 scale extends the overall dynamic range~\citep{nvidia2025nvfp4blog}.

\textbf{Post-Training Quantization (PTQ)}. PTQ is a simple and cost-effective method that requires no training~\citep{nagel2021white,haowu2020integerquantization,vanhoucke2011improving}. It involves a process called calibration that determines quantization parameters (e.g., scale factors and zero points) using a small representative dataset (calibration set). With PTQ, training and inference are completely decoupled, making it attractive for practitioners without access to original training pipelines or computational resources for fine-tuning.

The simplest PTQ method is max calibration, which sets the scale factor such that the maximum absolute value in the calibration data maps to the maximum representable value in the quantized format. Despite its simplicity, max calibration works surprisingly well in many cases. More sophisticated calibration methods have been proposed, including MSE-based calibration~\citep{jacob2018quantization}, KL divergence minimization~\citep{migacz2017}, and learnable clipping thresholds~\citep{choi2018pact}. More sophisticated PTQ methods have been proposed to minimize quantization error through advanced techniques. For example, AdaRound~\citep{nagel2020up} and BRECQ~\citep{li2021brecq} optimize weight rounding and block-wise reconstruction, respectively. For LLMs specifically, ZeroQuant~\citep{yao2022zeroquant} and GPTQ~\citep{frantar2023gptq} apply layer-wise calibration with efficient approximations. Other approaches focus on preserving sensitive information (AWQ~\citep{lin2023awq}) or optimizing quantization parameters via gradient descent (OmniQuant~\citep{shao2023omniquant}). A prominent direction is transformation-based PTQ, which applies reversible transformations (e.g., rotations, affine transformations) to suppress outliers and flatten distributions; examples include SmoothQuant~\citep{xiao2023smoothquant}, QuaRot~\citep{ashkboos2024quarot}, QuIP\#~\citep{tseng2024quipbetterllmquantization},  SpinQuant~\citep{liu2024spinquant}, and FlatQuant~\citep{sun2024flatquant}. Additionally, low-rank approximation methods like SVDQuant~\citep{svdquant2024} and EoRA~\citep{liu2025eora} have been proposed to mitigate quantization error by absorbing outliers or compensating for information loss. These advanced methods typically achieve better accuracy than simple max calibration, especially for lower bit-widths.

PTQ works quite well for FP8 on most LLMs. For NVFP4, PTQ also works well for very large models (See Appendix~\ref{sec:large_model_ptq_results}). However, PTQ often struggles with small models and sensitive tasks. Recent work has highlighted that common PTQ algorithms often fail to improve over baseline NVFP4 performance because the small block size neutralizes traditional outlier mitigation techniques~\citep{egiazarian2025bridging}. 


\subsection{Quantization-aware training}\label{sec:quantization_training}

Quantization-aware training (QAT)~\citep{jacob2018quantization} is used to recover inference accuracy after a model has been trained in full precision. Unlike native quantized training~\citep{deepseekv3,nvidia2025nvfp4pretraining}, which quantizes weights, activations and gradients to speedup both forward and backward passes, QAT only quantizes the weights and activations, thus only helping forward pass. Gradients remain in high precision to ensure stable convergence.

QAT fine-tunes the quantized model using the same loss function (e.g., cross-entropy for language modeling) and ideally the same training data as the original full-precision model. Modern LLMs typically undergo multi-stage post-training pipelines that include supervised fine-tuning (SFT) and reinforcement learning (RL) stages. For supervised training stages, QAT is straightforward to apply. For RL stages, the equivalent approach would be quantization-aware RL (QARL), which quantizes both the actor's forward pass and rollout generation. However, QARL remains less explored—existing work on quantized RL has focused on accelerating training throughput~\citep{huang2025qerl,yao2025flashrl} rather than post-hoc inference accuracy recovery. This motivates our investigation of QAD as an alternative for RL-trained models.

\subsection{Knowledge Distillation for Quantization}

Knowledge distillation (KD) ~\citep{hinton2015distilling} transfers knowledge from a teacher model to a student model by learning soft labels (e.g. probability distributions) from the teacher, typically using KL divergence loss for classification tasks. Theoretical work shows that soft labels provide a better estimate of true class probabilities with lower variance~\citep{menon2020distillation}, and that matching teacher distributions provides implicit regularization that accelerates convergence~\citep{phuong2021understanding}.

The combination of knowledge distillation and quantization has been studied extensively for CNNs. \citet{mishra2018apprentice} demonstrated that low-precision networks can be significantly improved through distillation from full-precision teachers. \citet{polino2018model} proposed quantized distillation, incorporating the distillation loss directly into the training of weight-quantized networks. \citet{kim2019qkd} introduced a three-phase training procedure to mitigate the strong regularization effect of KD on quantized models.

For large language models, \citet{liu2023llmqat} demonstrated that data-free distillation using model-generated data is highly effective, enabling quantization-aware training without access to the original training data. \citet{kim2023tsld} proposed token-scaled logit distillation, which reweights the per-token KL divergence based on teacher confidence to prevent overfitting. \citet{du2024bitdistiller} introduced BitDistiller for sub-4-bit LLMs, combining asymmetric quantization with a self-distillation objective that blends forward and reverse KL divergence.

Prior QAD work has primarily targeted integer quantization. This report demonstrates that simple KL divergence-based distillation is highly effective for NVFP4 inference accuracy recovery, particularly for LLMs trained through complex multi-stage post-training pipelines where replicating the original training is impractical and standard QAT risks degrading model capabilities.


\section{Quantization-Aware Distillation}
\label{sec:qad}
\subsection{Method Overview}

\begin{figure}[b!]
    \centering
    \includegraphics[width=0.7\textwidth]{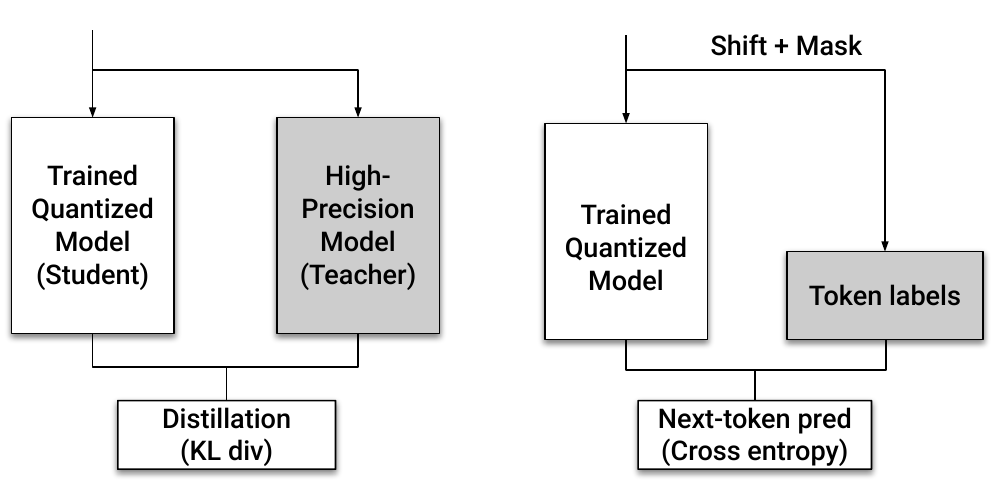}
    \caption{Comparison of quantization-aware training (QAT) and quantization-aware distillation (QAD). QAT trains with next-token prediction (cross-entropy) on target datasets, while QAD uses distillation loss (KL divergence) with the full-precision teacher model providing soft targets.}
    \label{fig:qad_qat}
\end{figure}

Quantization-aware distillation (QAD) uses the original high-precision model as a teacher to train the quantized model as a student using distillation loss. It differs from standard quantization-aware training (QAT), which trains the quantized model using the same task loss as the original model, as illustrated in Figure~\ref{fig:qad_qat}.

The key difference between QAD and QAT lies in the loss function:

\begin{itemize}
    \item \textbf{QAT}: Uses the same task-specific loss, e.g., next-token cross entropy for language modeling, as the original model training pipeline.
    \item \textbf{QAD}: Uses KL divergence between the high-precision teacher and the quantized student.
\end{itemize}

Formally, for a given input $x$ and vocabulary $V$, let $p_{\text{teacher}}(y|x)$ denote the output distribution from the full-precision teacher model and $p_{\text{student}}(y|x)$ denote the output distribution from the quantized student model. The QAD loss is the KL divergence between the teacher and student distributions:

\begin{equation}
\mathcal{L}_{\text{QAD}} = D_{\text{KL}}(p_{\text{teacher}} \| p_{\text{student}}) = \sum_{y \in V} p_{\text{teacher}}(y|x) \log \frac{p_{\text{teacher}}(y|x)}{p_{\text{student}}(y|x)}
\end{equation}





\begin{wraptable}{r}{0.5\textwidth}
    \centering
    \caption{QAD better aligns the model with BF16 baseline. Llama Nemotron Super V1 trained with $\sim$0.3B tokens sampled from its SFT dataset and evaluated on 5,000 held-out samples ($\sim$8M tokens).}\label{tab:llamanemotron_qad_qat_loss}
    \begin{tabular}{lcc}
    \toprule
    Methods & KL Divergence & Cross Entropy \\
    & (vs BF16) & (vs labels)\\
    \midrule
    BF16 & 0 & 0.408 \\
    QAT & 0.311 & 0.408 \\
    QAD & 0.004 & 0.416 \\
    \bottomrule
    \end{tabular}
\end{wraptable}

Table~\ref{tab:llamanemotron_qad_qat_loss} implies the key difference between QAD and QAT.
Remarkably, QAT achieves nearly the same cross-entropy loss as the BF16 baseline, which might suggest successful capability recovery. However, the KL divergence reveals a different story: QAD achieves nearly zero KL divergence, while QAT exhibits significant divergence from the teacher. This demonstrates that although QAT can match validation loss, it significantly changes the model's output distribution, effectively acting as an additional post-training stage. In contrast, QAD faithfully preserves the original BF16 model's output distribution.

\subsection{QAD for Post-Trained Models}

For single-stage post-training, we tested on Nemotron Nano 12B v2 VL~\citep{nvidia2025nvidianemotronnanov2} and found that QAT can match QAD performance (see Appendix~\ref{sec:llamanemotron_vl_results}). However, modern state-of-the-art LLMs typically undergo complex multi-stage pipelines involving SFT, RL, and model merging~\citep{bercovich2025llamanemotronefficientreasoningmodels,nvidia2025nvidianemotronnano2}. Applying QAT to such pipelines is challenging: it requires replicating each stage with quantized forward passes and reproducing model merging procedures between stages. A more practical approach is to run QAD or QAT as a single stage using a mixture of SFT data or model-generated data from RL prompts. In this simplified setting, we demonstrate that QAD consistently outperforms QAT, achieving near-BF16 accuracy regardless of the complexity of the original pipeline.

\begin{table}[ht]
    \centering
    \caption{Results on SFT-heavy models. Both QAD and QAT are trained with a mixture of the model's SFT data (math, coding, science, etc). QAD achieves near-BF16 accuracy and significantly outperforms QAT on reasoning tasks, especially on AIME25 and GPQA-D.}\label{tab:llamanemotronv1}
    \begin{tabular}{llcccc}
    \toprule
    Model & Method & MATH500 & AIME25 & GPQA-D  & IFEval-Instruction \\
    \midrule
    \multirow{4}{*}{\shortstack[l]{Llama Nemotron\\Super V1}} 
          & BF16         & 95.8   & 46.0  & 66.5 & 87.5 \\
          & NVFP4 PTQ    & 91.4   & 32.3  & 62.1 & 86.9 \\
          & NVFP4 QAT    & 94.3   & 41.5  & 63.3 & 87.2 \\
          & \textbf{NVFP4 QAD} & \textbf{94.6}   & \textbf{45.6}  & \textbf{64.5} & \textbf{87.8} \\
    \midrule
    \multirow{4}{*}{\shortstack[l]{Nemotron\\Nano V2}} 
          & BF16         & 97.8   & 71.1  & 64.0 & 90.3 \\
          & NVFP4 PTQ    & 97.2   & 69.8  & 59.0 & 89.8 \\
          & NVFP4 QAT    & 97.2    & 67.1  & 56.9 & 86.2 \\
          & \textbf{NVFP4 QAD} & \textbf{97.2} & \textbf{71.5}  & \textbf{62.7} & \textbf{89.3} \\
    \bottomrule
    \end{tabular}
\end{table}

\begin{table}[ht]
    \centering
    \caption{QAD results on RL-heavy models. For both models, NVFP4 QAT breaks the RL model's capabilities, while QAD successfully recovers near-BF16 performance, demonstrating the necessity of distillation for RL models.}\label{tab:rl_heavy_models}
    
    \textbf{(a) Nemotron 3 Nano}\protect\footnotemark
    
    \begin{tabular}{lcccccc}
    \toprule
    Method & AA-LCR & AIME25 & GPQA-D & LiveCodeBench-v5 & SciCode \\
    \midrule
    BF16         & 35.9 & 89.1 & 73.0 & 72.1 & 33.0 \\
    NVFP4 PTQ    & 31.3 & 85.0 & 71.6 & 68.9 & 30.5 \\
    \cmidrule{1-6}
    NVFP4 QAT    & 24.8 & 83.3 & 66.0 & 62.0 & 25.8 \\
    \textbf{NVFP4 QAD} & \textbf{34.3} & \textbf{87.9} & \textbf{72.7} & \textbf{68.9} & \textbf{32.3} \\ 

    \bottomrule
    \end{tabular}
    
    \vspace{1em}
    
    \textbf{(b) AceReason Nemotron 1.1 7B}
    
    \begin{tabular}{lccc}
    \toprule
    Method & AIME24 & AIME25 & LiveCodeBench-v6 \\
    \midrule
    BF16 Baseline & 73.0 & 63.5 & 54.3 \\
    NVFP4 PTQ & 69.4 & 58.7 & 52.0 \\
    \cmidrule{1-4}
    NVFP4 QAT & 62.1 & 46.1 & 45.9 \\
    \textbf{NVFP4 QAD} & \textbf{71.7} & \textbf{62.0} & \textbf{53.3} \\
    \bottomrule
    \end{tabular}
\end{table}
\footnotetext{Benchmark numbers differ from the model card, as different evaluation tools and settings are used.}


\newparagraph{SFT-Heavy Models.} We evaluate QAD on two SFT-heavy models that both undergo multi-stage post-training: Llama Nemotron Super V1 49B~\citep{bercovich2025llamanemotronefficientreasoningmodels} and Nemotron Nano 9B V2~\citep{nvidia2025nvidianemotronnano2}. Table~\ref{tab:llamanemotronv1} shows that QAD consistently outperforms QAT on challenging reasoning benchmarks across both models. For Llama Nemotron Super V1, QAD outperforms QAT particularly on AIME25 (+4.1\%) and GPQA Diamond (GPQA-D) (+1.2\%), recovering to near-BF16 performance. For Nemotron Nano 9B V2, QAD achieves near-BF16 performance and significantly outperforms QAT on AIME25 (+4.4\%) and GPQA-D (+5.8\%). These results demonstrate QAD's effectiveness on multi-stage trained models.

In addition, Nemotron Nano 12B v2 VL is a VLM model that undergoes a single SFT stage after pre-training. We reported the results in Appendix~\ref{sec:llamanemotron_vl_results}. 

\newparagraph{RL-Heavy Models} We evaluate QAD on two RL-heavy models: Nemotron 3 Nano 30B-A3B~\citep{nvidia2025nemotron3nanoopen}, a hybrid Mamba-Transformer model post-trained with multi-stage RL, and AceReason Nemotron 1.1 7B~\citep{chen2025acereasonnemotronadvancingmathcode,liu2025acereasonnemotron11advancingmath}, a Qwen2.5-based model~\citep{qwen2025qwen25technicalreport} specialized for math and code through RL. For RL-trained models, the RL training data typically contains only prompts, as the model generates responses during training. However, RL models are typically initialized from a cold-start SFT phase that teaches the base model to solve problems with chain-of-thought reasoning~\citep{deepseekai2025deepseekr1,chen2025acereasonnemotronadvancingmathcode,yang2025qwen3technicalreport,kimiteam2025kimik15,lingteam2025stepevolvesscalingreinforcement}. This cold-start SFT data provides a practical option for QAD and QAT training. Another option is to use generated data from RL prompts (see Section~\ref{sec:impact_of_training_data}) if cold-start SFT data is unavailable. In our experiments, we train Nemotron 3 Nano with a mixture of cold-start SFT data and RL-generated data, and AceReason with only cold-start SFT data.

However, using cold-start SFT data (or mixtures containing it) poses a fundamental challenge for QAT, as it can break the capabilities learned during RL training. Table~\ref{tab:rl_heavy_models} demonstrates this issue for both models: QAT significantly degrades performance across all benchmarks compared to PTQ, even when the training data includes RL-generated samples. In contrast, QAD successfully recovers near-BF16 performance, as it matches the teacher's output distribution rather than relearning from the data distribution. This demonstrates that distillation is essential for recovering accuracy in RL-trained models. An alternative approach would be to incorporate QAT into the RL training process itself, which remains an active research topic.

These results underscore QAD's key advantage for RL models: it avoids both the complexity of RL training and the risk of breaking learned capabilities, requiring only the full-precision teacher model. 


\subsection{Robustness to Incomplete Domain Coverage}

A key advantage of QAD is robustness to incomplete data coverage—the ability to recover accuracy even when the training data does not cover all domains or capabilities of the model.

\newparagraph{Cross-Domain Transfer on Multi-Domain Model.} AceReason Nemotron is trained on both math and code domains. Table~\ref{tab:AceReason_data_coverage} shows that QAD with partial data (math-only or code-only) nearly matches full data (math+code) results across all benchmarks. Remarkably, QAD trained with only code data recovers strong math performance, demonstrating effective cross-domain knowledge transfer through distillation.

\begin{table}[h!]
\centering
\caption{QAD on AceReason Nemotron 1.1 7B with partial domain coverage. QAD trained with math-only or coding-only data achieves near-full data performance, demonstrating cross-domain knowledge transfer.}\label{tab:AceReason_data_coverage}
\begin{tabular}{lccc}
\toprule
& AIME24 & AIME25 & LiveCodeBench-v6 \\
\midrule
BF16 Baseline & 73.0 & 63.5 & 54.3 \\
NVFP4 PTQ & 69.4 & 58.7 & 52.0 \\
\midrule
NVFP4 QAD (math only) & 71.0  & 61.7 & 53.1 \\
NVFP4 QAD (code only) & 71.0 & 62.0 & 53.3 \\
NVFP4 QAD (math+code) & 71.7 & 62.0 & 53.3 \\
\bottomrule
\end{tabular}
\end{table}

These results demonstrate that the teacher's output distributions encode implicit knowledge about all domains and capabilities, even when the input data comes from limited domains. By training the quantized student to match these distributions, QAD enables cross-domain knowledge transfer, allowing the student to approximate the teacher's behavior across domains not explicitly represented in the training data.

\subsection{Training and Evaluation Setup}

\textbf{Quantization Configuration.} For Llama Nemotron Super V1 and AceReason Nemotron, we quantize all GEMM layers to NVFP4. For Nemotron Nano 9B V2, a hybrid Mamba-Transformer architecture with 4 Transformer layers and 52 Mamba layers, we employ selective quantization: we keep attention layers in Transformer blocks and the first and last two layers at BF16 to maintain a better PTQ baseline. For Nemotron 3 Nano, a Mixture-of-Experts hybrid Mamba-Transformer with only 6 self-attention layers, we keep the 6 self-attention layers and their preceding Mamba-2 layers at BF16, quantize the remaining network to NVFP4, and quantize KV-Cache to FP8.

\textbf{Hyperparameters.} We use conservative learning rates: 1e-6 for Llama Nemotron Super V1 and Nemotron Nano 9B V2, and 1e-5 for AceReason Nemotron and Nemotron 3 Nano. The softmax temperature is set to $T=1$ for both teacher and student to precisely match the teacher's output distribution. Batch sizes and sequence lengths are kept similar to those used in the original post-training.

\textbf{Data.} QAD requires significantly less data than the original post-training. The exact amount depends on the model size and task complexity. We report the amount of data required for convergence. 
\begin{itemize}
    \item Llama Nemotron Super V1 49B: $\sim$0.3 billion tokens
    \item Nemotron Nano 9B V2: $\sim$6 billion tokens
    \item Nemotron Nano 12B V2 VL: $\sim$0.5 billion tokens
    \item Nemotron 3 Nano 30B-A3B: $\sim$2.5 billion tokens 
    \item AceReason Nemotron 7B: $\sim$0.8 billion tokens  
\end{itemize}

\textbf{Evaluation.} For each experiment, we evaluate the top 10 checkpoints with the lowest validation loss and select the one that performs best on average across evaluation benchmarks. For Llama Nemotron Super V1, Nemotron Nano 9B V2, and AceReason Nemotron, we report results using multiple sampling runs per problem: 48 runs for AIME 2024~\citep{aime24} and AIME 2025~\citep{aime25}, 12 runs for LiveCodeBench~\citep{livecodebench}, 20 runs for GPQA Diamond (GPQA-D)~\citep{rein2023gpqa}, and 5 runs for IFEval~\citep{zhou2023instructionfollowing}. All evaluations use temperature $T=0.6$ and top-$p=0.95$ for sampling. For Nemotron 3 Nano, we report average results across: 16 runs for AIME 2025, 8 runs for LiveCodeBench-v5, 8 runs for GPQA-D, and 5 runs for AA-LCR~\citep{artificialanalysis2025lcr} and SciCode~\citep{tian2024scicode}. For SciCode, we report subtask accuracy. All evaluations use temperature $T=1.0$ and top-$p=1.0$ for sampling.


\section{Ablation study}\label{sec:ablation}

This section provides detailed ablation studies of QAD, specifically focusing on the impact of training data quality and learning rate sensitivity.

\begin{table}[h!]
\centering
\caption{Impact of training data on QAD for AceReason Nemotron 1.1 7B. We test different data sources: (1) \textbf{SFT data}: original cold-start SFT data used during training; (2) \textbf{Generated from RL prompts}: BF16-generated samples from RL prompts; (3) \textbf{Generated from RL prompts (correct only)}: filtered to include only correct solutions; (4) \textbf{Generated from BOS token}: data generated by providing a single initial token following~\citet{liu2023llmqat}; (5) \textbf{Random tokens}: completely random token sequences. QAD shows remarkable robustness across all data sources. All experiments use the same amount of data.}\label{tab:rl_data_quality}

\vspace{0.5em}

\begin{tabular}{lccc}
\toprule
Training data & AIME24 & AIME25 & LiveCodeBench-v6 \\
\midrule
BF16 Baseline & 73.0 & 63.5 & 54.3 \\
NVFP4 PTQ & 69.4 & 58.7 & 52.0 \\
\midrule
\textbf{SFT data} & \textbf{71.7} & \textbf{62.0} & \textbf{53.3}  \\
\textbf{Generated from RL prompts} & \textbf{71.9} & \textbf{61.3} & \textbf{52.6} \\
Generated from RL prompts (correct only) & 70.5 & 61.6 & 52.3 \\
Generated from BOS token & 70.1 & 60.9 & 52.4 \\
Random tokens & 68.6 & 60.0 & 51.7 \\
\bottomrule
\end{tabular}

\end{table}

\subsection{Training Data Quality}
\label{sec:impact_of_training_data}

An important practical question of QAD is the sensitivity to data quality and source. We conduct ablation studies on AceReason Nemotron to understand the impact of different data sources. Similar robustness is observed for Nemotron 3 Nano (see Appendix~\ref{sec:nemotron3nano_data_quality}).

Table~\ref{tab:rl_data_quality} shows QAD's performance with different data sources for AceReason Nemotron. Cold-start SFT data and BF16-generated data from RL prompts both achieve near-BF16 performance, demonstrating that synthetic data is highly effective for QAD. Interestingly, using all generated samples (including incorrect ones) performs better than using only correct samples, suggesting that incorrect generations also provide useful information for distillation. Even with completely random tokens, QAD maintains comparable performance to the PTQ baseline without breaking the model, demonstrating remarkable stability.

For Nemotron 3 Nano, we observe similar robustness across different data sources (SFT data, RL-generated data, and their mixtures), with all achieving comparable performance (see Appendix~\ref{sec:nemotron3nano_data_quality} for details). These results demonstrate that QAD is remarkably robust to data source and quality. 
\begin{table}[b]
    \centering
    \caption{Learning rate sensitivity of QAD for AceReason Nemotron 1.1 7B (RL-heavy) and Nemotron Nano 9B V2 (SFT-heavy). For AceReason, the optimal LR (1e-5) is notably higher than standard RL rates, while for Nano V2 9B, increasing the learning rate above the original SFT LR (1e-6) degrades performance or causes unstable training.}
    \label{tab:Ablation_lr_sensitivity}
    \begin{tabular}{lcccc}
    \toprule
    Model & Learning Rate & AIME24 & AIME 25 & LiveCodeBench \\
    \midrule
     & 1e-6 & 70.8 & 61.0 & 52.6 \\
     \multirow{2}{*}{AceReason Nemotron 1.1 7B} & 5e-6 & 71.0 & 60.9 & 53.2  \\
     & \textbf{1e-5} & \textbf{71.7} & \textbf{62.0} & \textbf{53.3}  \\
     & 1e-4 & 72.4 & 61.8 & 53.0  \\
    \midrule
    \multirow{2}{*}{Nemotron Nano 9B V2} 
     & \textbf{1e-6} & \textbf{80.4} & \textbf{71.5} & \textbf{67.8}  \\
     & 5e-6 & 80.0 & 71.0 & 66.8  \\
     & 1e-5 & 80.8 & 69.4 & 67.4  \\
     & 1e-4 & 78.8 & 65.2 & 64.0  \\ 
    \bottomrule
    \end{tabular}
\end{table}

\begin{table}[t]
    \centering
    \caption{Learning rate sensitivity for Nemotron Nano 12B v2 VL (SFT-trained). The model achieves optimal performance at LR of 2e-6, which is 10× lower than its original SFT LR (2e-5).}
    \label{tab:Ablation_lr_vlm_sensitivity}
    \begin{tabular}{ccccccc}
        \toprule
         Learning Rate & AI2D & ChartQA & DocVQA & InfoVQA & OCRBench & TextVQA \\
        \midrule
         1e-4 & 67.0 & 76.0 & 75.0 & 47.6  & 685 & 70.6 \\
         2e-5 & 85.3 & 87.6 & 91.6 & 72.2 & 820 & 82.8 \\
         \textbf{2e-6} & \textbf{87.1} & \textbf{89.7} & \textbf{94.0} & \textbf{78.9} & \textbf{857} & \textbf{84.7} \\
        \bottomrule
    \end{tabular}%
\end{table}

\subsection{Learning Rate}
\label{sec:hyperparameter_sensitivity}
QAD requires careful learning rate selection, with different optimal ranges depending on the original training. Overall, we recommend a learning rate between 1e-5 to 1e-6.

For SFT-trained models, we find that using learning rates at or below the original post-training learning rate works best. As shown in Tables~\ref{tab:Ablation_lr_sensitivity} and~\ref{tab:Ablation_lr_vlm_sensitivity}, Nemotron Nano 9B V2 achieves optimal performance at 1e-6 (matching its original SFT learning rate), while Nemotron Nano 12B v2 VL performs best at 2e-6 (10× lower than its original SFT learning rate of 2e-5). Higher learning rates degrade performance or even diverges, likely because these models are already well-converged on the SFT data distribution after post-training.

For RL-trained models, the situation is notably different. Since the final RL stage shifts the model away from the cold-start SFT data distribution, QAD benefits from larger learning rates. Table~\ref{tab:Ablation_lr_sensitivity} shows that for AceReason Nemotron, the optimal QAD learning rate is 1e-5, which is substantially larger than typical RL learning rates ($\sim$1e-6)~\citep{shao2024deepseekmath,deepscaler2025,chen2025acereasonnemotronadvancingmathcode}.



\begin{table}[h!]
    \centering
    \caption{KL divergence vs. MSE on different models. KL divergence consistently outperforms MSE.}\label{tab:Ablation_kl_div_mse}
    \begin{tabular}{lccccc}
    \toprule
    Model & Loss & GPQA-D & AIME24 & AIME25 & LiveCodeBench \\
    \midrule
    \multirow{2}{*}{AceReason Nemotron 1.1 7B} & KL-Div & / & 71.7 & 62.0 & 53.3  \\
     & MSE & / & 71.7 & 60.1 & 52.4  \\
    \midrule
    \multirow{2}{*}{Nemotron Nano 9B V2} & KL-Div & 62.7 & 80.4 & 71.5 &  67.8 \\
     & MSE & 60.3 & 80.0 & 71.5 & 66.7  \\
    \bottomrule 
    \end{tabular}
\end{table}

\subsection{Additional Choices}
\newparagraph{KL Divergence vs. MSE.} We use KL divergence as the distillation loss for QAD, which is the standard choice for matching probability distributions. While other distance metrics such as MSE on logits could theoretically be used, Table~\ref{tab:Ablation_kl_div_mse} shows that KL divergence consistently outperforms MSE across benchmarks. This is likely because KL divergence is better suited to measure distributional differences and provides better gradients for probability matching.

\begin{wraptable}{r}{0.5\textwidth}
    \centering
    \caption{Results of Nemotron Nano 9B V2 NVFP4 distilled by different teachers. Using the original model as teacher outperforms using a larger teacher. }\label{tab:larger_teacher}
    \begin{tabular}{lcccc}
    \toprule
     Teacher & AIME24 & AIME25 & LiveCodeBench \\
    \midrule
     9B BF16 & 80.4 & 71.5 & 67.8  \\
     12B BF16 & 80.2 & 69.8 & 66.7  \\
    \bottomrule
    \end{tabular}
\end{wraptable}

\newparagraph{Using a Larger Teacher.} Unlike traditional knowledge distillation where a larger teacher transfers knowledge to a smaller student, QAD uses the original BF16 model as the teacher to recover its exact distribution. While using a larger teacher from the same model family is feasible, as they typically trained with similar data and techniques. We tested this on Nemotron Nano 9B V2 using both the original 9B BF16 teacher and a larger 12B BF16 teacher. Table~\ref{tab:larger_teacher} shows that the 9B teacher outperforms the 12B teacher, One potential reason is that adapting to a different distribution requires more triaining data compared to typical QAD. For efficient accuracy recovery, we still recommend using the original model as the teacher.

\section{Conclusion}

This technical report presents quantization-aware distillation (QAD) as a practical and effective method for recovering inference accuracy of LLMs and VLMs quantized to NVFP4 format. Through experiments on Nemotron Nano, Nemotron Nano VL, Llama Nemotron Super, and AceReason Nemotron, we show that QAD reliably brings NVFP4 models back to near-BF16 accuracy across a wide range of tasks, including models trained with complex SFT, RL, and model merging pipelines where replicating the original training procedure is impractical for standard QAT.

Our ablations further demonstrate that QAD is robust to data coverage and quality: it can leverage partial-domain or synthetic data, and remains stable even when trained on random tokens. Combined with modest data and compute requirements compared to original post-training, these properties make QAD a practical default for NVFP4 accuracy recovery when PTQ alone is insufficient. The NVFP4 QAD checkpoints and code are available at the links provided in the abstract, enabling practitioners to adopt these techniques in real deployments.


\newpage
\bibliographystyle{plainnat}
\bibliography{references}

\newpage
\appendix

\section{Llama Nemotron VL results}
\label{sec:llamanemotron_vl_results}

Nemotron Nano 12B v2 VL is a VLM model that undergoes a single SFT stage after pre-training. unlike other post-trained LLMs, QAT for this model achieves comparable accuracy to QAD. We believe this is attributed to the simple training pipeline and also the small accuracy drop from PTQ.

\begin{table}[h!]
    \centering
    \caption{Accuracy comparison of different methods on Llama Nemotron Nano 12B v2 VL with AI2D~\citep{kembhavi2016diagramworthdozenimages}, ChartQA~\citep{masry2022chartqabenchmarkquestionanswering}, DocVQA~\citep{mathew2021docvqadatasetvqadocument}, InfoVQA~\citep{mathew2021infographicvqa}, OCRBench~\citep{Liu_2024}, and TextVQA~\citep{singh2019towards}}
    \label{tab:Ablation_vlm_ptq_qat_qad}
    \begin{tabular}{ccccccc}
        \toprule
         Method & AI2D & ChartQA & DocVQA & InfoVQA & OCRBench & TextVQA \\
        \midrule
         Baseline & 87.3 & 89.7 & 94.3 & 79.3  & 855 & 85.2 \\
         PTQ & 86.8 & 89.6 & 93.8 & 78.2  & 850 & 84.8 \\
         QAT & 86.5 & 89.8 & 93.7 & 78.3 & 848 & 84.8 \\
         \textbf{QAD} & \textbf{86.7} & \textbf{89.4} & \textbf{93.9} & \textbf{78.4} & \textbf{858} & \textbf{85.2} \\
        \bottomrule
    \end{tabular}%
\end{table}

\section{Nemotron 3 Nano Data Quality Ablation}
\label{sec:nemotron3nano_data_quality}

For Nemotron 3 Nano, we test QAD with three different data sources: (1) cold-start SFT data only, (2) BF16-generated data from RL prompts only, and (3) a mixture of both. Table~\ref{tab:nemotron3nano_data_quality} shows that all three data sources achieve similar performance, with the SFT+RL mixture performing slightly better.

\begin{table}[h!]
\centering
\caption{Impact of training data on QAD for Nemotron 3 Nano 30B-A3B. All three data sources achieve comparable performance, demonstrating QAD's robustness to data composition.}\label{tab:nemotron3nano_data_quality}

\vspace{0.5em}

\begin{tabular}{lccccc}
    \toprule
    Training data & AA-LCR & AIME25 & GPQA-D & LiveCodeBench-v5 & SciCode \\
    \midrule
    BF16 Baseline & 35.9 & 89.1 & 73.0 & 72.1 & 33.0  \\
    NVFP4 PTQ & 31.3 & 85.0 & 71.6 & 68.9 & 30.5 \\
    \cmidrule{1-6}
    SFT data & 32.6 & 86.0 & 72.7 & 70.0 & 31.7 \\
    Generated from RL prompts & 34.0 & 82.7 & 73.9 & 70.4 & 33.1 \\
    {SFT+RL generations mixture} & {34.3} & {87.9} & {72.7} & {68.9} & {32.3} \\ 
    \bottomrule
\end{tabular}

\end{table}

\section{PTQ for Large Models}
\label{sec:large_model_ptq_results}

An empirical finding is that larger LLMs are more robust to quantization. In Table~\ref{tab:large_model_ptq}, we show one example each for NVIDIA trained models and community models. More results and quantized checkpoints can be found under  \href{https://huggingface.co/collections/nvidia/inference-optimized-checkpoints-with-model-optimizer}{HuggingFace - NVIDIA collections}.

\begin{table}[ht]
    \centering
    \caption{PTQ results on large models (hundreds of billions of parameters). Large models are robust to NVFP4 PTQ, achieving near-original accuracy without any fine-tuning.}\label{tab:large_model_ptq}
    \begin{tabular}{llcccc}
    \toprule
    Model & Method & MATH500 & AIME24 & GPQA-D & GSM8K\\
    \midrule
    \multirow{2}{*}{\shortstack[l]{Llama Nemotron\\Ultra V1 (253B)}} 
          & BF16         & 96.6   & 75.0  & 75.7 & / \\
          & NVFP4 PTQ    & 96.2   & 76.0  & 74.8 & / \\
    \midrule
    \multirow{2}{*}{\shortstack[l]{DeepSeek R1\\(671B)}} 
          & Official FP8   & 95.4   & 80.0  & 69.7 & 96.3 \\
          & NVFP4 PTQ      & 94.2   & 80.0  & 69.2 & 96.1 \\
    \bottomrule
    \end{tabular}
\end{table}

\section{Comparison of QAT/QAD with native quantized training}

\begin{figure}[h!]
    \centering
    \includegraphics[width=0.9\textwidth]{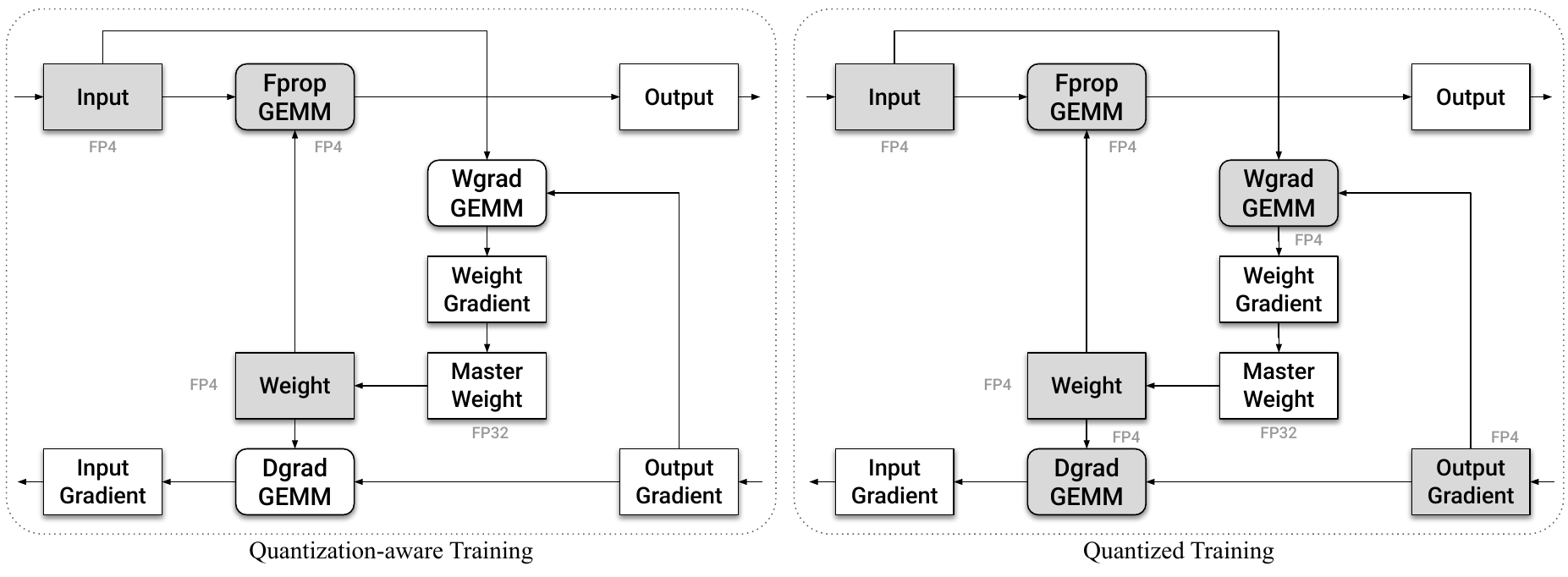}
    \caption{Comparison of quantization-aware training (QAT) and native quantized training. QAT only quantizes the forward pass for inference recovery, while native quantized training quantizes all three GEMMs (Fprop, Wgrad, Dgrad) to reduce training cost. QAD has a similar compute graph as QAT.}
    \label{fig:qat_native}
\end{figure}

\textbf{Native quantized training} traces its origins to mixed precision training (FP16/BF16), with both approaches sharing the primary goal of reducing computational cost during training. Native quantized training is primarily used during pretraining, where training is compute-bound since batch sizes can be made arbitrarily large. The core idea is to quantize three GEMMs: forward propagation (Fprop), weight gradient (Wgrad), and data gradient (Dgrad). To perform these GEMMs in low precision, all three of their inputs (activations, weights, and output gradients) must be quantized. Examples include DeepSeek V3's FP8 training~\citep{deepseekv3} and recent NVFP4 and MXFP4 pretraining~\citep{nvidia2025nvfp4pretraining,chmiel2025fp4,chen2025tetrajet,rouhani2023microscaling}. These methods focus on reducing the cost of pretraining frontier models from scratch. 

QAT/QAD select the quantization targets similarly, as both of them quantize weights and/or activations while leaving gradients in high precision. As a result, the two GEMMs (Wgrad and Dgrad) in backward pass cannot be calculated in lower precision. Figure~\ref{fig:qat_native} illustrates the key difference between QAT and quantized training.

\end{document}